%% file: arxiv.tex
\documentclass[10pt,twocolumn,letterpaper]{article}

\usepackage[pagenumbers]{cvpr} % To force page numbers, e.g. for an arXiv version
\usepackage{booktabs,tabularx}
\usepackage{xcolor}
\usepackage{amssymb}
\usepackage{graphicx}
\usepackage{cuted}
\usepackage{caption}
\usepackage{placeins}

\DeclareRobustCommand{\robotmark}[1]{\textcolor[HTML]{#1}{\rule{0.9ex}{0.9ex}}}
\input{preamble}
\definecolor{cvprblue}{rgb}{0.21,0.49,0.74}
\usepackage[pagebackref,breaklinks,colorlinks,allcolors=cvprblue]{hyperref}

\def\confName{CVPR}
\def\confYear{2027}

\title{Wrist-Camera Video Diffusion as Generalizable Robot Policies}

\author{
    Jingyuan Liu$^{1,2}$ \qquad
    Yongwei Chen$^{1,3}$ \qquad
    Yishan Yang$^{1,4}$ \\
    Zexin He$^{1,3}$ \qquad
    Xingang Pan$^{3}$\thanks{Corresponding authors.} \qquad
    Zhi Hou$^{1}$\footnotemark[1] \\[2pt]
    $^{1}$ACE Robotics \qquad
    $^{2}$Fudan University \\
    $^{3}$Nanyang Technological University \qquad
    $^{4}$Southeast University
}
\newcommand{\methodname}{Wrist Video Policy}
\newcommand{\methodnameshort}{WVP}

\begin{document}
\maketitle

\begin{strip}
    \centering
    \includegraphics[width=\textwidth]{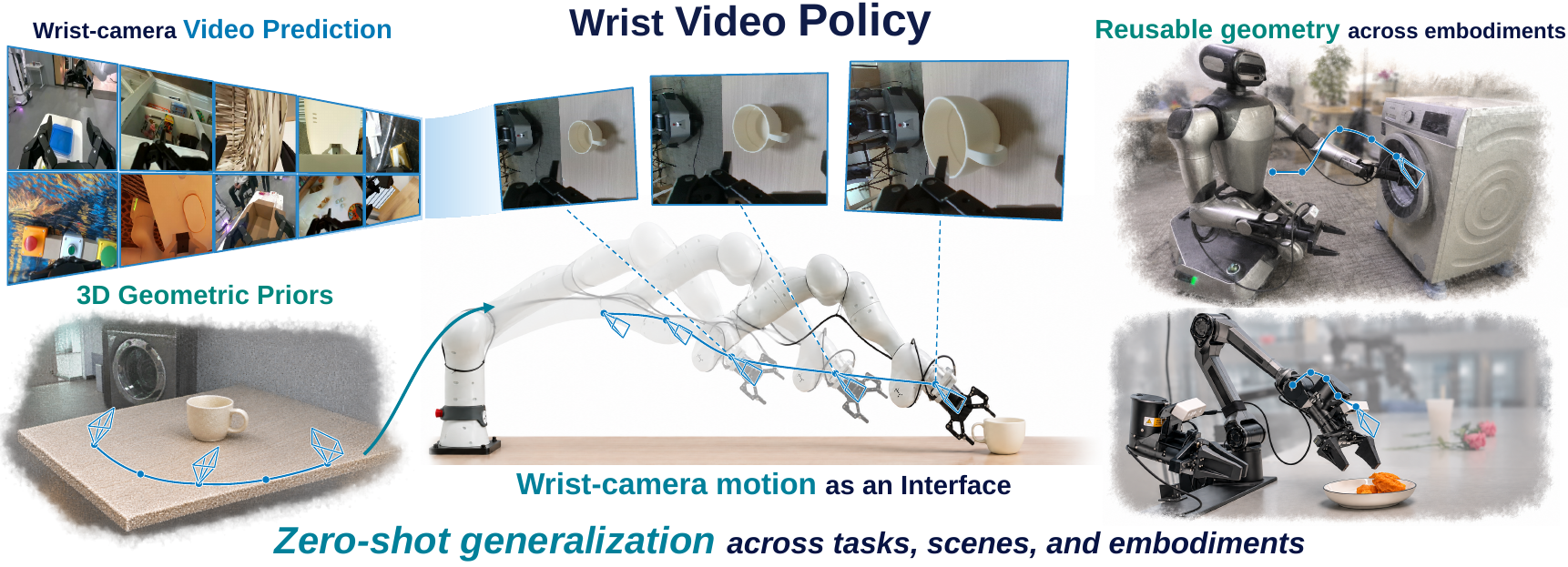}
    \captionof{figure}{
        \textbf{Video prediction meets wrist-camera motion estimation.} WVP uses wrist-camera motion as a shared action representation, connecting task knowledge from pretrained video models with geometric priors from 3D models. Video prediction captures how a task unfolds, while geometric motion estimation translates predicted views into executable actions across tasks, scenes, and embodiments.
    }
    \label{fig:intro}
\end{strip}

\input{sec/0_abstract}    
\input{sec/1_intro}
\input{sec/2_related}
\input{sec/3_preliminaries}
\input{sec/4_methods}
\input{sec/5_experiments}
\input{sec/6_conclusion}
\section*{LLM Usage Statement}
We used OpenAI's ChatGPT and Codex to help revise the manuscript, organize related work and citations, refine code and figures, and clarify the presentation of our datasets, experimental design, results, and conclusions. We also used RoboJudge-9B to evaluate generated videos, as described in Sec.~\ref{sec:component}. The authors take full responsibility for all submitted content.
{
    \small
    \bibliographystyle{ieeenat_fullname}
    \bibliography{references}
}

\input{sec/X_suppl}

\end{document}

%% file: sec/0_abstract.tex
\begin{abstract}
% A generalist manipulation policy should perform new tasks, operate in unfamiliar scenes, and transfer skills across robot embodiments. We present \methodname{}~(\methodnameshort{}), a closed-loop manipulation framework that represents robot motion through wrist-camera trajectories. By estimating the wrist-camera trajectory, we could recover the corresponding end-effector poses needed for execution through a fixed hand-eye transformation. The framework combines a video generator with a pose estimator. Given a language instruction and current observations, the generator predicts how the wrist views will evolve. To recover the camera motion from these predictions, we adapt a pretrained 3D model using wrist-camera videos and trajectories from five public robot datasets. Model stitching connects the two models in the generator's latent space, allowing the estimator to recover wrist-camera trajectories without first decoding the video into RGB frames. The resulting pose estimator transfers to Franka, PiperX, and Galaxea R1 Pro without further fine-tuning. Across tabletop and laundry-room settings, \methodnameshort{} generalizes to tasks ranging from precise grasping and stacking to handling deformable objects. Video demonstrations are included in the supplementary material.

A generalist manipulation policy should perform new tasks, operate in unfamiliar scenes, and transfer skills across robot embodiments. We present \methodname{}~(\methodnameshort{}), a closed-loop manipulation framework that learns how manipulation unfolds through video prediction and recovers executable motion through geometric estimation. Our key idea is to represent robot motion through wrist-camera trajectories, which determine the corresponding end-effector poses through a fixed hand-eye transformation. We pair a video generator that predicts future wrist views with a pose estimator adapted from a pretrained 3D model to recover the corresponding camera trajectories. Model stitching maps the generator's predicted latents directly into the estimator's intermediate feature space, bypassing RGB decoding. This decomposition allows task knowledge to expand through broad human and robot video data while reusing geometric motion estimation across tasks and embodiments. The resulting pose estimator transfers across Franka Panda, PiPER-X, and Galaxea R1 Pro without further fine-tuning. Across tabletop and laundry-room settings, \methodnameshort{} generalizes to tasks ranging from precise grasping and inserting to handling deformable objects, achieving a success rate of over 50\% in zero-shot out-of-distribution evaluations. Our project page is at~\href{https://wrist-video-policy.github.io}{https://wrist-video-policy.github.io}.
\end{abstract}

%% file: sec/1_intro.tex
\section{Introduction}
\label{sec:intro}

%%%%%%%%%%%%%%% VERSION-1 
% Scaling manipulation data has enabled robot policies to perform new tasks, operate in unfamiliar environments, and transfer skills across embodiments~\cite{black2025pi05,physicalintelligence2026pi07}. These advances draw on several complementary ways of scaling training data. One approach combines demonstrations and autonomous rollouts from multiple robot platforms with egocentric human videos and multimodal web data~\cite{physicalintelligence2026pi07}. Others scale human manipulation data by recording physical interactions with wearable devices~\cite{generalist2026gen1} or by tracking hand motion in large collections of first-person videos to obtain action supervision~\cite{dyna2026dyna2}. These policies can generalize across robot embodiments and perform dexterous skills, for example by folding a T-shirt on a robot that was never trained to fold laundry~\cite{physicalintelligence2026pi07}. Despite this progress, most policies still learn to act through imitation, using motions recorded in robot demonstrations or inferred from human videos as supervision~\cite{physicalintelligence2026pi07,dyna2026dyna2}. This raises a practical question. Can we build on pretrained visual models to achieve zero-shot manipulation across tasks and embodiments without further fine-tuning on action-labeled demonstrations?

Generalization in robotic manipulation requires policies to perform new tasks and transfer familiar skills across scenes and robot embodiments~\cite{chen2025robotwin}. Vision-language-action (VLA) models combine pretrained visual and linguistic knowledge with robot demonstrations, enabling manipulation across diverse tasks, environments, and embodiments~\cite{brohan2023rt,kevin2025pi,intelligence2026pi}. Scaling these capabilities through action-labeled demonstrations, however, creates an expensive coverage problem, as the training data must capture a broad range of tasks, environments, and robot motions. Video-based approaches, including world-action models (WAMs), offer a way to learn much of this task and interaction knowledge from a broader supply of human and robot videos, without requiring corresponding action labels for video pretraining~\cite{hongtao2023unleashing,zhu2025unified}. Recent advances demonstrate the potential of this approach through transfer from human videos and zero-shot execution of new tasks~\cite{robotics2026dyna,yanjiang2026video}. Yet learning to faithfully translate visual predictions into executable motion still typically requires action-labeled robot trajectories~\cite{ye2026world,yanjiang2026video}.
Even with this reliance on action labels, separating task knowledge from motion execution opens a promising path to scalable robot learning.

Building on this separation, we aim to achieve zero-shot transfer across tasks, scenes, and embodiments without additional action-labeled demonstrations for deployment. A natural foundation for this goal is the 3D geometry of manipulation. Different robots may need different joint commands to approach and manipulate an object, but their motion can be described in the same way across embodiments by how far the gripper moves and how it is oriented relative to the object. Recent policies incorporate 3D scene features into action prediction~\cite{tsungwei2024d,qu2025spatialvla}, but still rely on robot demonstrations to learn the mapping from geometry to actions, limiting zero-shot generalization across tasks, scenes, and embodiments. Pretrained 3D models, on the other hand, suggest a more direct role for geometry. They can already estimate camera poses from images across diverse scenes and viewpoints~\cite{jianyuanvggt,lin2025depth,wang2026vggtomega}. Treating motion execution as a well-defined geometric estimation problem could extend these pretrained capabilities to robot control and, together with video prediction, provide a scalable path towards learning manipulation in 3D visual space.

A wrist-mounted camera makes this idea concrete by connecting video prediction with geometric estimation. Because it is rigidly attached to the end effector, their poses are related by a fixed hand-eye transformation. We can therefore plan the robot's motion by predicting future wrist-camera views, estimating the camera trajectory, and converting it into end-effector poses for execution. The prediction target remains the motion of the observing camera regardless of which robot carries it, allowing the same estimator to transfer across embodiments through each robot's hand-eye calibration. This lets task knowledge expand through video prediction while reusing the geometric mechanism that translates those predictions into actions.

To this end, we introduce \textbf{\methodname{}~(\methodnameshort{})}, a policy that connects pretrained video and 3D models through wrist-camera motion. Given a language instruction and current observations, the generator predicts how wrist views will evolve, drawing on pretrained video models that generalize across manipulation tasks and scenes~\cite{lester2026turning,yanjiang2026video}. The pose estimator builds on a pretrained 3D model~\cite{jianyuanvggt}, whose geometric priors we adapt to robot control through fine-tuning on public wrist-camera videos and trajectories. Once adapted, it is reused across tasks and embodiments without further fine-tuning. We then connect the two components through model stitching~\cite{go2025texttod}, mapping the predicted video latents directly into the estimator's intermediate feature space. This bypasses RGB decoding and yields action trajectories that more closely follow the visual plan.

We evaluate \methodnameshort{} on real robots across tasks, scenes, and embodiments. Trained solely on public robot datasets, the pose estimator transfers across all three settings without further fine-tuning. Paired with a video generator, it enables zero-shot manipulation with a success rate of over 50\% in our out-of-distribution tests. Across tabletop and laundry-room settings, the system performs tasks ranging from pick-and-place and stacking to wiping, inserting, retrieving clothes, and opening articulated fixtures. Its capabilities span fine-grained manipulation of small rigid objects, handling of deformable garments and towels, and precise stacking of transparent tableware~(Fig.~\ref{fig:demo}). In summary, our contributions are threefold.
\begin{itemize}
    \item We represent robot actions using wrist-camera motion, formulating action prediction as estimating the camera poses of future wrist-camera views.
    \item We present a pose estimator built on a pretrained 3D model, and use model stitching to estimate wrist-camera motion directly from a video generator's latents.
    \item We demonstrate that \methodnameshort{} generalizes across tasks, scenes, and embodiments in the real world, achieving a success rate of over 50\% in zero-shot evaluations.
\end{itemize}

%% file: sec/2_related.tex
\section{Related Work}

\subsection{Generalization in Robotic Manipulation}

Vision-language-action models combine pretrained visual and linguistic knowledge with robot demonstrations from diverse tasks and embodiments~\cite{brohan2023rt,jin2024openvla,black2025pi}. Recent approaches scale vision-language models for embodied perception and reasoning~\cite{team2026ace,team2025robobrain}, express robot motions in natural language to benefit from language pretraining~\cite{zha2026lap}, or unify action spaces for joint training across embodiments~\cite{haoqi2026qwenrobotmanip}. Lower-cost data collection methods further support training across diverse settings~\cite{cheng2024universal,songming2026rdt, genesis2026gene265}. These advances enable robots to manipulate unfamiliar objects~\cite{brohan2023rt}, perform household tasks in unseen homes~\cite{kevin2025pi}, and transfer familiar skills to new task--embodiment combinations~\cite{intelligence2026pi,luo2026beingh}. However, existing low-cost data collection methods typically rely on shared end-effector configurations, limiting their applicability across heterogeneous robot platforms. Moreover, despite reduced collection costs, scaling action-labeled demonstrations across embodiments remains expensive, motivating the use of human and robot videos without corresponding action annotations. 
% Even with lower collection costs, obtaining action-labeled demonstrations at scale remains expensive, motivating the use of human and robot videos without corresponding action labels.

World-action models learn from these videos by predicting future observations from earlier frames~\cite{hongtao2023unleashing,chilam2024gr}. Scaling this pretraining has improved transfer from human videos to robot manipulation~\cite{homanga2024genact,robotics2026dyna,heng2026openwam} and enabled zero-shot execution of new tasks~\cite{yilun2023learning,kevin2023zeroshot,yanjiang2026video}, but learning actual robot commands still typically requires action-labeled trajectories~\cite{ye2026world,yanjiang2026video}. To reduce this dependence, some methods learn latent actions from unlabeled videos~\cite{ye2024latent,wenyao2026disentangled} or train an inverse dynamics model independently of the video backbone using task-agnostic robot exploration~\cite{hengkai2025anypos,lester2026turning,zhang2025generative}. These approaches reduce the need for task-specific demonstrations, but do not fully eliminate the reliance on robot-action data~\cite{hengkai2025anypos}. Geometric methods instead derive robot trajectories from hand motion, object poses, or 3D object flow estimated from video~\cite{pochenlearning,junbang2024dreamitate,shivansh2025robotic,chen2025large,karthik2025dreamflow,hanzhi2025vidbot}. However, estimating robot end-effector poses from object trajectories or hand poses often relies on accurate depth estimates and a clear view of the relevant motion, and errors in reconstructing that motion can propagate to the resulting robot trajectory. \methodnameshort{} instead represents robot actions using the motion of the wrist camera itself, and draws on the geometric priors of a pretrained 3D model to estimate that motion from video predictions. Each robot's hand-eye calibration maps the estimated trajectory to end-effector targets, enabling zero-shot manipulation across embodiments.

\subsection{Model Stitching}

Video generators encode how a scene will change in their latent representations, before any RGB frames are decoded. Model stitching enables reuse of these representations by learning a map from one model's features to an intermediate layer of another. Originally used to study whether networks learn compatible representations~\cite{karelunderstanding,yamini2021revisiting,csiszrik2021similarity}, stitching has since been used to combine pretrained components across architectures and modalities~\cite{xingyi2022deep,zizheng2023stitchable,jaisidh2025almost,zheda2026revisiting}. VIST3A~\cite{go2025texttod}, for example, maps video latents into a pretrained 3D reconstruction network to generate scene geometry without video decoding. In robot control, video latents can also serve as inputs to action decoders that predict the commands needed to carry out a visual plan~\cite{yucheng2024video,jonas2025mimicvideo,junbang2025video, yanjiang2026video}. We connect these two ideas by stitching a video generator to a pose estimator built on a pretrained 3D model~\cite{jianyuanvggt}, significantly improving performance in real-world manipulation.

%% file: sec/3_preliminaries.tex
\section{Preliminaries}

\label{sec:preliminaries}

\begin{figure*}[t]
    \centering
    \begin{subfigure}[t]{0.61\textwidth}
        \centering
        \includegraphics[width=\linewidth]{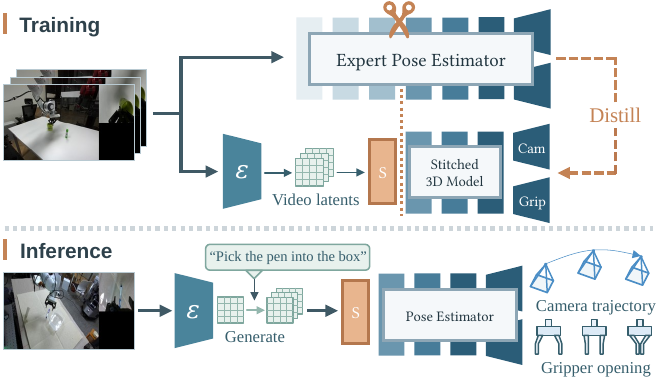}
        \caption{Training and inference.}
        \label{fig:overview}
    \end{subfigure}\hfill
    \begin{subfigure}[t]{0.37\textwidth}
        \centering
        \includegraphics[width=\linewidth]{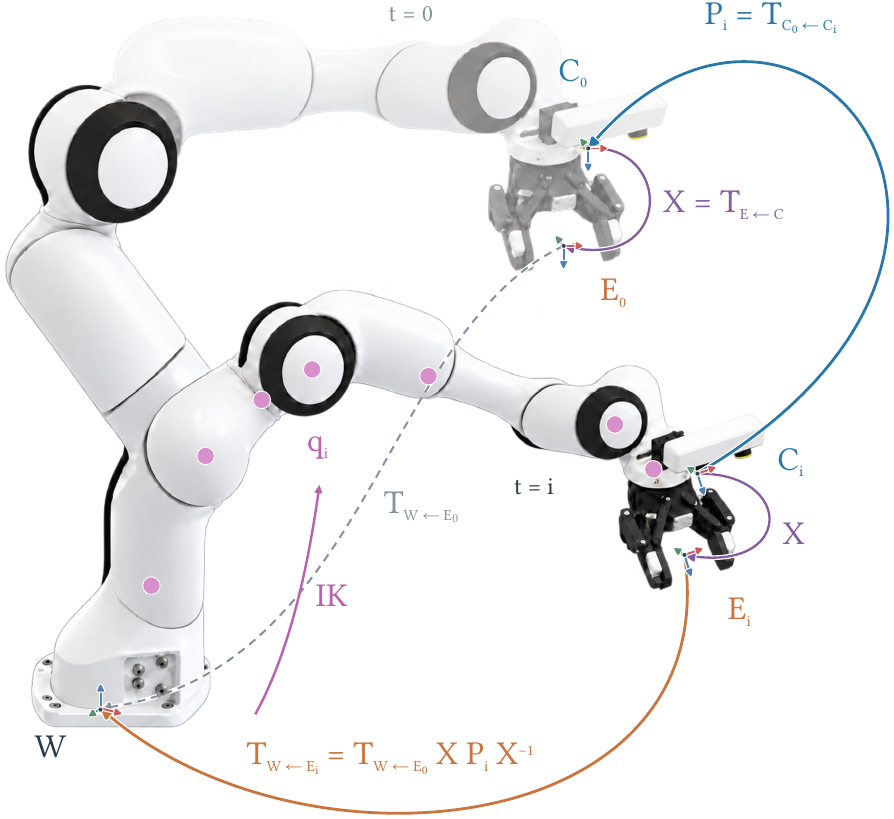}
        \caption{From camera motion to robot actions.}
        \label{fig:pose-calib}
    \end{subfigure}

    \caption{\textbf{From video predictions to robot actions with \methodnameshort{}.} (a) A frozen RGB pose expert supervises the stitched estimator's camera-pose and gripper predictions. The VAE encoder $\varepsilon$ produces video latents, and the adapter $S$ maps them into the pose estimator's feature space. At inference, the estimator recovers motion directly from predicted latents without RGB decoding. (b) Relative camera motion $P_i$ and hand-eye calibration $X$ determine end-effector targets from the current pose. Inverse kinematics (IK) converts these targets into joint configurations $q_i$. $C$, $E$, and $W$ denote the camera, end-effector, and reference frames, respectively. Labels are defined in Sec.~\ref{sec:preliminaries}.}
    \label{fig:intro}
\end{figure*}

We aim to develop a closed-loop robotic manipulation system that turns video predictions into executable actions. A wrist camera is related to the end effector by a fixed transformation, so its trajectory provides a geometric description of end-effector motion. We build on this relationship by combining a general video model with a 3D pose estimator. Given a language instruction and current observations, the video model predicts future wrist-camera observations, from which the pose estimator recovers the camera trajectory. This casts robot motion generation as a wrist-camera pose estimation problem. The recovered trajectory is converted into robot actions, and the system executes the action segment before re-planning from new observations.

To describe this conversion~(Fig.~\ref{fig:pose-calib}), let $T_{A\leftarrow B}$ denote a rigid transformation that maps coordinates from frame $B$ to frame $A$. For one arm, let $P_i=T_{C_0\leftarrow C_i}\in\mathrm{SE}(3)$ denote the future camera pose relative to the current camera frame $C_0$, and let $X=T_{E\leftarrow C}$ be the fixed camera-to-end-effector transformation obtained through hand-eye calibration. Given the current end-effector pose $T_{W\leftarrow E_0}$ in the execution reference frame $W$, the future target is
\begin{equation}
    T_{W\leftarrow E_i}
    =
    T_{W\leftarrow E_0}\,X\,P_i\,X^{-1}.
    \label{eq:handeye}
\end{equation}
Inverse kinematics maps this target to a joint configuration, which is executed together with the predicted scalar gripper value. For bimanual manipulation, the same conversion is applied independently to each arm using its own hand-eye calibration and the end-effector pose.

%% file: sec/4_methods.tex
\section{Methods}

\methodnameshort{} connects task knowledge from pretrained video models with geometric motion estimation from a pretrained 3D model through wrist-camera trajectories. We first adapt the 3D model on public robot videos paired with wrist-camera trajectories and gripper states, yielding an RGB pose estimator that can be reused across tasks, scenes, and embodiments~(Sec.~\ref{sec:expert_training}). We then use model stitching to map the generator’s predicted video latents into the estimator’s intermediate feature space, enabling motion estimation without RGB decoding~(Sec.~\ref{sec:model_stitching}). At deployment, the estimated camera trajectories are converted into end-effector targets through the robot’s hand-eye calibration and executed with the predicted gripper states in a closed loop~(Sec.~\ref{sec:inference}). 

\subsection{Training the Expert Pose Estimator}
\label{sec:expert_training}

To learn a shared geometric motion estimator across embodiments, we curate wrist-camera RGB videos paired with camera trajectories and gripper states from AgiBot~\cite{contributors2025agibot}, DROID~\cite{khazatsky2024droid}, BEHAVIOR-1K~\cite{li2024behaviork}, InternData-A1~\cite{tian2025interndataa}, and RH20T~\cite{fang2023rht}. Camera poses are obtained from dataset annotations or reconstructed from robot states and camera calibrations, with base motion taken into account when expressing poses in a moving robot base. We unify camera-axis conventions across datasets and express each wrist-camera trajectory relative to its first frame, using world-to-camera transforms as supervision following VGGT~\cite{jianyuanvggt}. Specifically, given camera poses $C_i=T_{W\leftarrow C_i}$ in a fixed reference frame $W$, the supervision target is $Q_i=C_i^{-1}C_0=P_i^{-1}$, where $C_0$ is the first-frame pose and $P_i$ is the camera-to-first-frame transformation defined in Sec.~\ref{sec:preliminaries}. We retain translations in meters instead of applying VGGT’s scene-scale normalization, preserving the metric scale required for precise execution.

General 3D model backbones are normally followed by multi-head output tailored for multiple functions~\cite{jianyuanvggt, lin2025depth,wang2026vggtomega}. Our formulation uses camera extrinsics to recover end-effector motion, allowing us to focus the model on camera-pose estimation. We use VGGT as the default backbone, retaining its camera head to predict camera translation and rotation while removing intrinsic prediction and the depth, point-map, and tracking heads. To complement camera motion with grasp-and-release control, we add a two-layer MLP head that predicts continuous gripper states from the shared backbone features. Gripper labels from each dataset are normalized to $[0, 1]$.

Starting from VGGT’s pretrained backbone and camera head, we jointly fine-tune both components with the gripper head on wrist-camera RGB videos, adapting the shared representation through camera-pose and gripper supervision. We optimize VGGT’s translation and rotation losses together with a smooth $L_1$ loss for gripper prediction:
\begin{equation}
\mathcal{L}
=
w_1\bigl(
\mathcal{L}_{\mathrm{trans}}
+
\mathcal{L}_{\mathrm{rot}}
\bigr)
+
w_2\mathcal{L}_{\mathrm{gripper}},
\label{eq:training_loss}
\end{equation}
where $\mathcal{L}_{\mathrm{trans}}$, $\mathcal{L}_{\mathrm{rot}}$, and $\mathcal{L}_{\mathrm{gripper}}$ supervise camera translation, rotation, and gripper status, respectively. The weights balance pose estimation and gripper regression.

\subsection{Model Stitching}
\label{sec:model_stitching}

Applying the RGB pose estimator to the generator’s predictions requires first decoding video latents into RGB frames. However, applying the RGB expert to decoded predictions can yield inaccurate camera trajectories and degrade task execution~(Sec.~\ref{sec:component}). Inspired by VIST3A~\cite{go2025texttod}, we use model stitching to map video latents directly into the pose estimator’s intermediate feature space, allowing its geometric estimation capabilities to operate on the predicted latents without RGB decoding.

We choose the stitching layer by measuring how accurately a linear map can reconstruct the expert’s intermediate features from video latents. For $N$ paired feature samples, let $B\in\mathbb{R}^{N\times d_z}$ denote video latents from the generator’s frozen VAE encoder and $A_k\in\mathbb{R}^{N\times d_k}$ the corresponding expert features at candidate backbone layer $k$, where $d_z$ and $d_k$ are their respective feature dimensions. We fit a linear map $S_k\in\mathbb{R}^{d_z\times d_k}$ for each candidate layer and select the layer with the lowest reconstruction error averaged over samples and feature dimensions:
\begin{equation}
\begin{aligned}
S_k^\star
&=
\arg\min_{S_k}
\left\|B S_k-A_k\right\|_F^2, \\
k^\star
&=
\arg\min_k
\frac{1}{N d_k}
\left\|B S_k^\star-A_k\right\|_F^2,
\end{aligned}
\label{eq:stitching_selection}
\end{equation}
where $\|\cdot\|_F$ denotes the Frobenius norm.

We initialize a trainable adapter $S$ with the selected linear map $S_{k^\star}^\star$, replacing the expert’s first $k^\star$ layers to accept video latents as input. The student’s remaining backbone layers and camera and gripper heads are initialized from the trained expert, yielding the stitched pose estimator
\begin{equation}
F_{\mathrm{stitch}}(z) = G_{k^\star+1:L}\bigl(S(z)\bigr),
\label{eq:stitched_pose_estimator}
\end{equation}
where $z$ denotes wrist-view video latents and $G_{k^\star+1:L}$ comprises the remaining layers of the $L$-layer backbone followed by the camera and gripper heads.

To transfer the expert’s motion estimation capabilities to latent inputs, we distill the frozen RGB pose estimator into the stitched student using the training videos from Sec.~\ref{sec:expert_training}. We encode the training videos with the frozen VAE encoder, following the generator's view layout and preprocessing. The student takes these latents as input and learns to match the camera-pose and gripper predictions produced by the RGB expert at corresponding time steps, requiring no additional action labels. We jointly optimize the adapter, the retained backbone, and both prediction heads to adapt the pretrained representation to latent inputs while preserving the expert's geometric capabilities~(Fig.~\ref{fig:overview}, top).

\subsection{Closed-Loop Inference}
\label{sec:inference}

At each planning step, the video generator conditions on the current observations, encoded by its VAE encoder $\varepsilon$, and the language instruction to predict video latents for $M$ future time steps~(Fig.~\ref{fig:overview}, bottom).
The stitched pose estimator takes the current and predicted wrist-view latents and estimates camera trajectories and gripper states for each arm. We invert each predicted world-to-camera transform to obtain the relative camera pose $\hat{P}_i=\hat{Q}_i^{-1}$, then apply the robot’s hand-eye calibration through Eq.~\eqref{eq:handeye} to obtain the end-effector target. Inverse kinematics converts these targets into joint configurations. The robot executes the first $K$ actions with the corresponding gripper commands before replanning from updated observations. The execution horizon $K$ balances feedback responsiveness and inference cost. A shorter execution horizon enables earlier correction of execution errors at the cost of more frequent inference.

%% file: sec/5_experiments.tex
\section{Experiments}

\begin{figure*}[t]
    \centering
    \includegraphics[width=\textwidth]{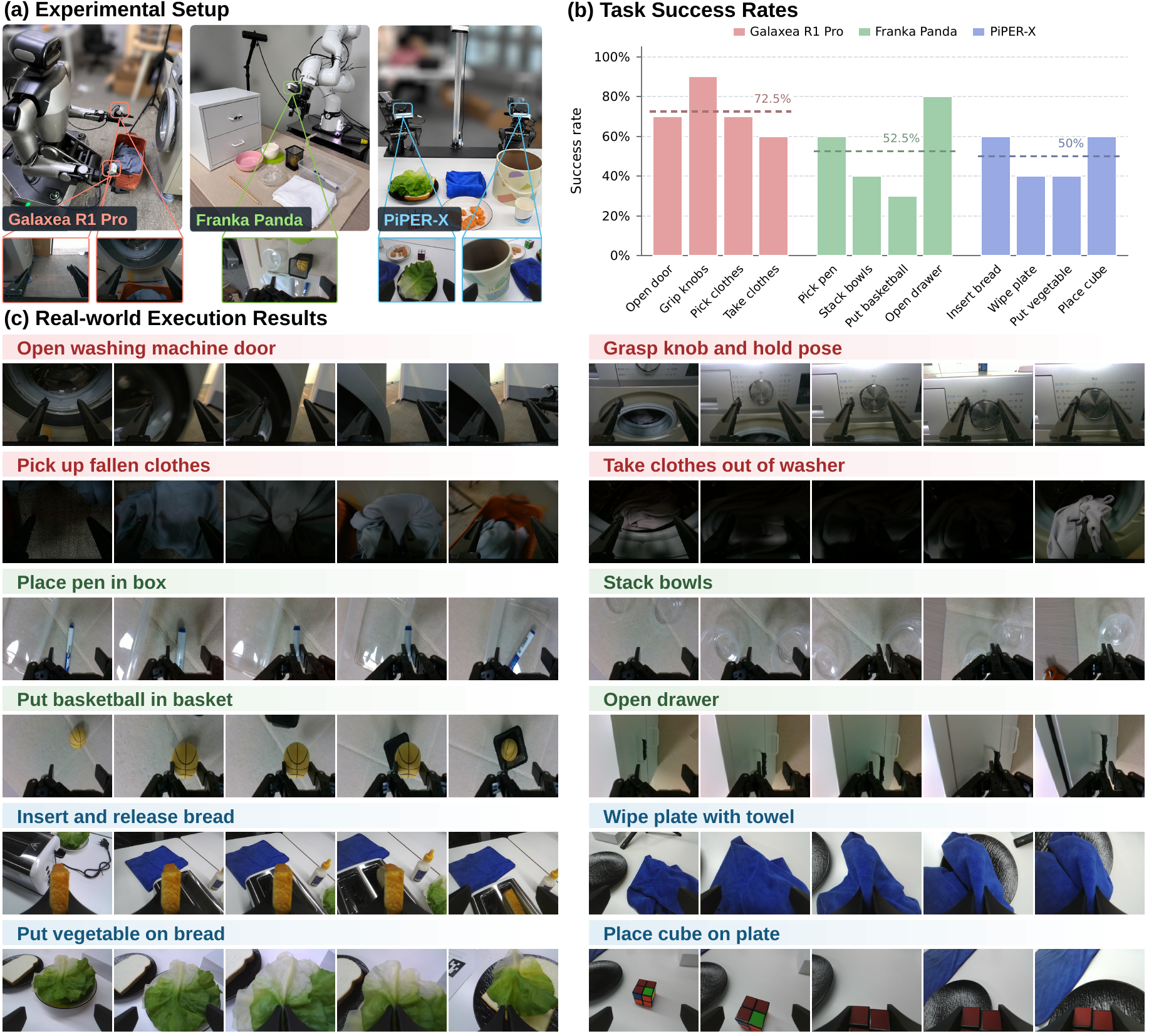}
    \caption{\textbf{Real-world manipulation across three embodiments.} The same pose estimator is reused on all three platforms without further fine-tuning.
    (a) Robot platforms and camera configurations, with insets showing wrist-mounted camera views.
    (b) Success rates over 10 trials per task. Dashed lines indicate the mean across the four tasks on each platform.
    (c) Wrist-camera views from representative executions, progressing from left to right.
    Task-label colors identify the platform:
    \robotmark{B95151}~Galaxea R1 Pro,
    \robotmark{428657}~Franka Panda, and
    \robotmark{437BA8}~PiPER-X.
    }
    \label{fig:demo}
\end{figure*}

We evaluate \methodnameshort{}’s ability to turn video predictions into executable actions across tasks, scenes, and embodiments. We first examine real-world manipulation with a pose estimator reused across these settings without further fine-tuning, including demonstrations of recovery from execution disturbances~(Sec.~\ref{sec:real_world}). We then compare WVP with existing robot-control approaches to assess its task success rates~(Sec.~\ref{sec:comparison}). Finally, we analyze how video prediction, adaptation of the pretrained 3D model, and latent-space stitching affect the accuracy of motion estimation and the success of task execution~(Sec.~\ref{sec:component}).

\subsection{Real-World Manipulation}
\label{sec:real_world}

\paragraph{Manipulation across tasks and embodiments.}

We evaluate \methodnameshort{} on Franka Panda, PiPER-X, and Galaxea R1 Pro, with their camera configurations shown in Fig.~\ref{fig:demo}a.\footnote{Hereafter, we refer to Franka Panda and Galaxea R1 Pro as Franka and Galaxea, respectively.} Unless otherwise specified, all experiments use the same stitched VGGT pose estimator trained as described in Sec.~\ref{sec:model_stitching}, without further fine-tuning. The video generator is used directly on Franka and PiPER-X, and adapted using only RGB recordings of the target tasks on Galaxea. The evaluation covers 12 tasks across tabletop manipulation on Franka and PiPER-X and laundry-room manipulation on Galaxea, including pick-and-place, opening, inserting, grasping, stacking, and wiping. These tasks pose different challenges. Grasping a pen and a basketball requires different gripper states. Opening a washing machine door or a drawer requires handling articulated fixtures. Handling clothes and towels requires maintaining a valid grasp as they deform.

Across 10 trials per task, WVP achieves mean success rates of 72.5\% on Galaxea, 52.5\% on Franka, and 50.0\% on PiPER-X~(Fig.~\ref{fig:demo}b). These results show that a pose estimator trained on five public robot datasets can support manipulation across all three platforms without further fine-tuning, translating video predictions into executable motion across different tasks and scenes. The executions in Fig.~\ref{fig:demo}c further illustrate WVP’s behavior under challenging visual and manipulation conditions, as detailed below.
\begin{itemize}
    \item \textbf{Manipulation in low light.} Galaxea retrieves clothes from the floor and inside the washing machine, where the wrist views become dark during approach. Despite limited visibility, it grasps and retrieves the garments.

    \item \textbf{Fine-grained manipulation.} Galaxea aligns its gripper with the washing machine’s recessed handle, while Franka engages the drawer handle before pulling. Franka also grasps a slender pen just above the tabletop, demonstrating precise gripper positioning.

    \item \textbf{Transparent-object manipulation.} Franka picks up and stacks transparent bowls despite reflections and overlapping rims that obscure their boundaries in the wrist view.
\end{itemize}
Across these conditions, the same pose estimator converts predicted wrist views into executable motion, supporting its reuse across diverse manipulation requirements. Please refer to the supplementary material for more videos.

\paragraph{Recovery from execution deviations.}

\begin{figure}[t]
    \centering
    \includegraphics[width=\linewidth]{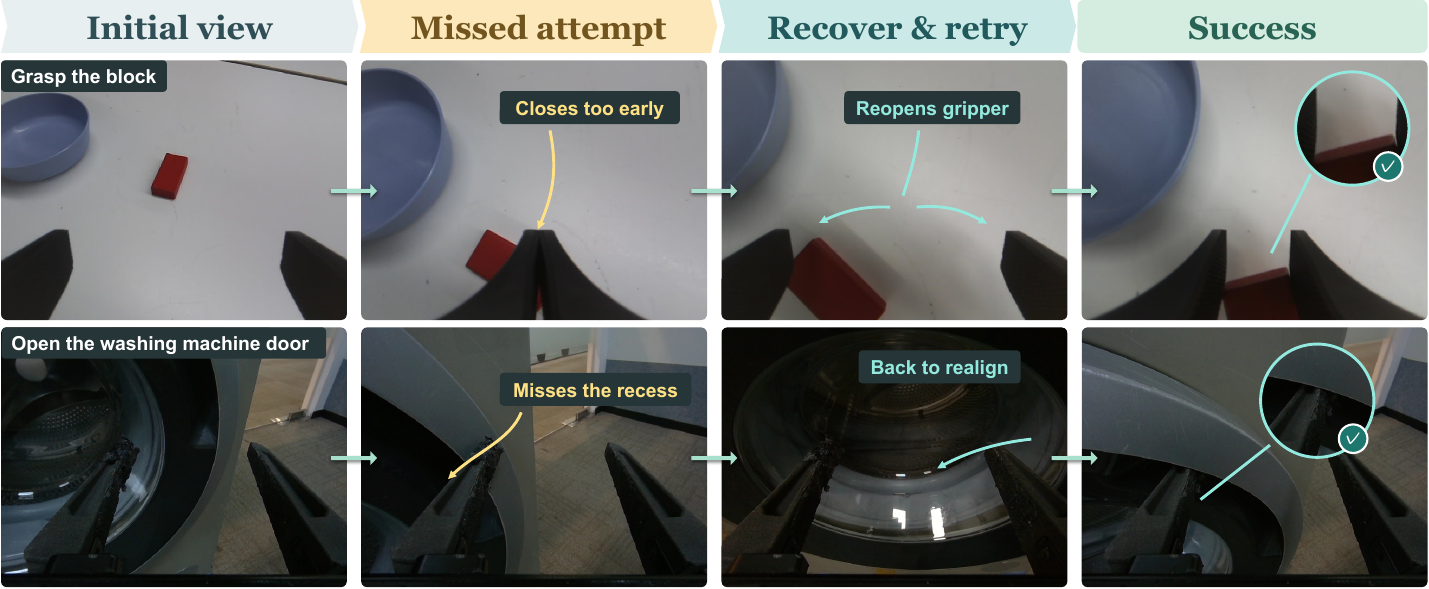}
    \caption{\textbf{Recovery from execution deviations.}
    PiPER-X retries a grasp after displacing the block~(top), and Galaxea realigns its gripper after missing the recessed door handle~(bottom).}
    \label{fig:recovery}
\end{figure}

Execution errors can leave the robot or manipulated objects in states that the original visual plan did not anticipate. By conditioning subsequent video predictions on new observations, \methodnameshort{} can adjust its approach and retry from these states. In Fig.~\ref{fig:recovery}~(top), PiPER-X closes its gripper too early and displaces the block instead of grasping it. From the updated wrist view, the generator predicts a new approach, and the robot reopens its gripper and grasps the block at its new position. In Fig.~\ref{fig:recovery}~(bottom), Galaxea initially misses the washing machine’s recessed handle, then backs away, realigns its gripper, and opens the door on the next attempt. Together, these examples show how closed-loop feedback and the video model’s ability to replan from out-of-distribution states enable WVP to correct execution deviations.

\subsection{Comparison to Alternative Policies}
\label{sec:comparison}

\begin{figure}[t]
    \centering
    \includegraphics[width=\linewidth]{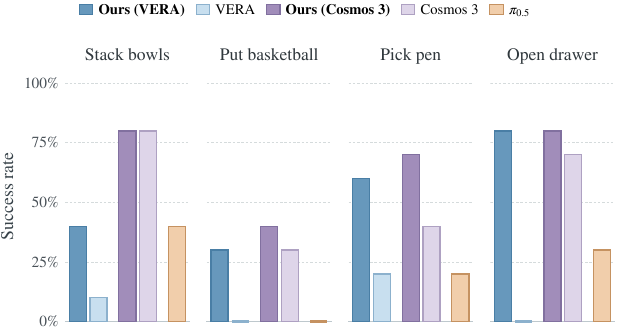}
    \caption{\textbf{Policy comparison on four Franka manipulation tasks.} Each bar reports the success rate over 10 trials under the same evaluation conditions. Ours (VERA) and Ours (Cosmos3) pair the VERA-DROID and Cosmos3-Nano-DROID video backbones, respectively, with our stitched wrist-camera pose estimator.}
    \label{fig:franka-policy-comparison}
\end{figure}

To compare how well different policies generalize across manipulation tasks, we compare \methodnameshort{} with baselines on the four Franka tasks from Sec.~\ref{sec:real_world}. All methods use the same robot setup, camera configuration, and evaluation protocol. We pair our stitched pose estimator with the VERA-DROID video generator~\cite{lester2026turning} without deployment-specific fine-tuning of either component, and use DROID-fine-tuned checkpoints for the baselines.
As shown in Fig.~\ref{fig:franka-policy-comparison},
\methodnameshort{} substantially outperforms VERA and $\pi_{0.5}$~\cite{kevin2025pi}, while achieving performance comparable to Cosmos3-Nano-DROID~\cite{agarwal2026cosmos}. With the same VERA-DROID video generator, our stitched pose estimator converts predicted wrist-camera motion into robot actions and achieves higher success rates than VERA's Jacobian IDM on all four tasks. This result supports camera motion as an effective, reusable action representation that connects video prediction to robot control through shared geometric motion estimation.

We further test the applicability of this representation and the stitching approach by pairing \methodnameshort{} with the Cosmos3-Nano-DROID video backbone. Following Sec.~\ref{sec:model_stitching}, we retrain the stitched pose estimator for the latent representation of the underlying Wan2.2 model, without further adaptation to the DROID-specific video checkpoint or the target embodiment. With the Cosmos3-Nano-DROID video backbone, WVP again outperforms the model’s original action head, extending the advantage of the camera-motion representation to a second video generator. This pairing also achieves higher success rates than our VERA-based policy, demonstrating that the stitching approach accommodates different video generators and allows their predictions to improve downstream execution. Together, these results support a scalable design in which wrist-camera motion serves as a shared action representation across video generators, and suggest that improvements in video prediction can translate into higher manipulation success rates through geometric motion estimation.

\subsection{Component Analysis}
\label{sec:component}

We evaluate how video generation, pose-estimation backbones, pretrained initialization, pose supervision, and model stitching affect motion estimation and task execution.

\paragraph{Video generator selection.}

We evaluate six video generators on 256 clips spanning 12 tasks collected with PiPER Dual Arm. Following RoboJudge~\cite{siyuan2026robojudge}, we assess Physical Adherence (PA), measuring physical plausibility, and Instruction Alignment (IA), measuring progress toward the instructed task. As shown in Tab.~\ref{tab:video_generators}, VERA scores highest on PA and Cosmos3 on IA. However, qualitative examples in the supplementary material show that VERA often makes little progress toward the goal, whereas Cosmos3 introduces substantial errors in scene content. Though VPP2 ranks second on both metrics, it achieves the best balance between physical faithfulness and task progression on PiPER.

To assess how these prediction differences translate into execution, we pair each generator with its corresponding trained pose estimator and conduct ten trials of \textit{put the vegetable on the bread}. The VPP2 pairing succeeds in 4/10 trials, compared with 1/10 for VERA and 0/10 for the other generators. Nevertheless, generator performance varies across deployment settings. We use VPP2 on PiPER-X and VERA-DROID on Franka, where it outperforms VPP2 in our trials. Neither generator performs well on the Galaxea laundry tasks without adaptation, so we fine-tune VERA using only RGB recordings from that setting.

\begin{table}[t]
    \centering
    \small
    \caption{\textbf{Video generation and real-world execution on PiPER.}
    PA and IA are mean scores (1--5) for the paired wrist views. VERA denotes VERA-Omni, VPP2 denotes its Stage 1 video model, and Cosmos3 denotes the Cosmos3-Super video model. }
    \label{tab:video_generators}
    \begin{tabular}{@{}lccc@{}}
        \toprule
        Generator & PA $\uparrow$ & IA $\uparrow$ & Success $\uparrow$ \\
        \midrule
        Cosmos Predict2.5~\cite{arslan2025world} & 1.0000 & 1.2422 & 0/10 \\
        LVP~\cite{chen2025large}  & 1.4688 & 1.7344 & 0/10 \\
        Wan2.1~\cite{wan2025wan}   & 2.0625 & 1.7070 & 0/10 \\
        Cosmos3~\cite{agarwal2026cosmos} & 2.2188 & \textbf{2.6367} & 0/10 \\
        VERA~\cite{lester2026turning}  & \textbf{3.5313} & 1.8672 & 1/10 \\
        VPP2~\cite{yanjiang2026video}  & 2.7422 & 2.1563 & \textbf{4/10} \\
        \bottomrule
    \end{tabular}
\end{table}

\paragraph{Pose estimator selection.}

\begin{figure*}[t]
    \centering
    \includegraphics[width=\textwidth]{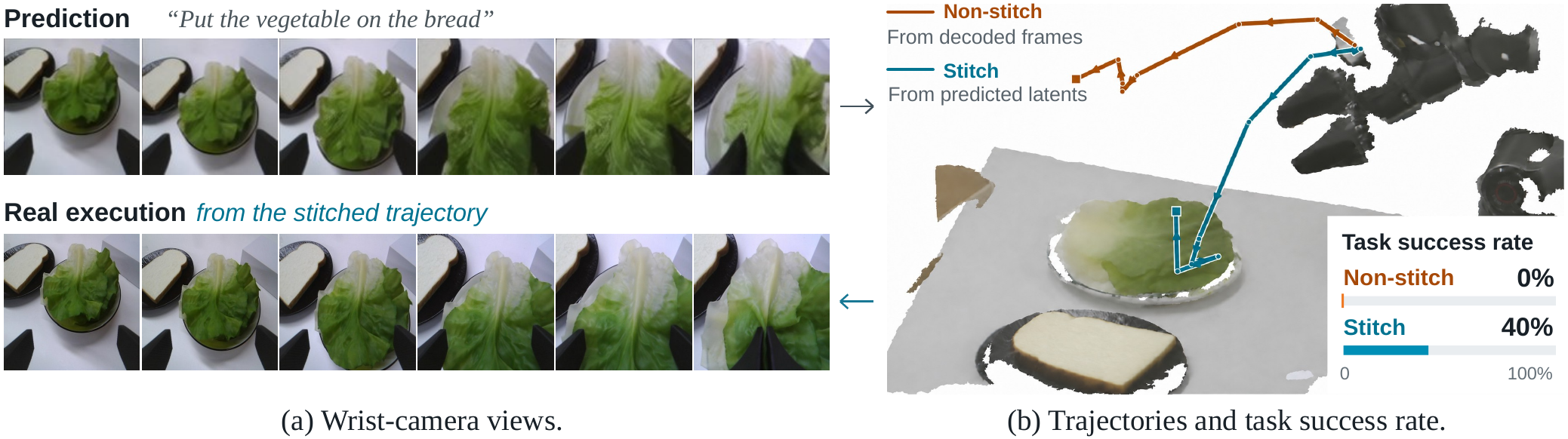}
    \caption{\textbf{Effect of model stitching on motion estimation and execution.}
    (a) Predicted wrist-camera views (top) and real execution using the stitched trajectory (bottom) for \textit{put the vegetable on the bread} on PiPER-X.
    (b) Camera trajectories recovered from the same video prediction. The RGB expert processes decoded frames, while the stitched estimator directly processes predicted latents.
    }
    \label{fig:stitch-ablation}
\end{figure*}

\begin{table}[t]
    \centering
    \caption{\textbf{Offline pose estimation on 1,000 held-out Galaxea episodes.} Translation and rotation are evaluated using root mean squared error (RMSE), and gripper states using mean absolute error (MAE) on the normalized $[0,1]$ scale.}
    \label{tab:pose_estimators}
    \small
    \setlength{\tabcolsep}{3pt}
    \begin{tabular}{@{}lccccc@{}}
        \toprule
        & \multicolumn{2}{c}{Original}
        & \multicolumn{3}{c}{Fine-tuned} \\
        \cmidrule(lr){2-3} \cmidrule(lr){4-6}
        Model
        & \shortstack{Trans.\\(mm) $\downarrow$}
        & \shortstack{Rot.\\($^\circ$) $\downarrow$}
        & \shortstack{Trans.\\(mm) $\downarrow$}
        & \shortstack{Rot.\\($^\circ$) $\downarrow$}
        & \shortstack{Gripper\\MAE $\downarrow$} \\
        \midrule
        Fast3R~\cite{yang2025fastr}
        & 129.32 & 34.81 & 95.98 & 13.00 & 0.297 \\
        DUSt3R~\cite{shuzhe2023dustr}
        & 137.69 & 23.80 & 92.37 & 12.67 & 0.238 \\
        MonST3R~\cite{zhang2024monstr}
        & \textbf{116.18} & 16.76 & 92.32 & 12.93 & 0.286 \\
        DA3~\cite{lin2025depth}
        & 229.41 & 13.05 & 76.08 & 9.42 & 0.315 \\
        VGGT-$\Omega$~\cite{wang2026vggtomega}
        & 208.79 & \textbf{9.27} & 69.52 & 7.23 & 0.236 \\
        VGGT~\cite{jianyuanvggt}
        & 191.19 & 10.92 & \textbf{65.29} & \textbf{6.84}
        & \textbf{0.203} \\
        \bottomrule
    \end{tabular}
\end{table}

We evaluate six pose estimators on wrist-camera videos from 1,000 held-out Galaxea episodes. The laundry-room environment is absent from all five fine-tuning datasets, allowing us to assess transfer to an unseen scene. We convert predicted camera poses to end-effector poses and compute errors against the recorded poses. As shown in Tab.~\ref{tab:pose_estimators}, all six pretrained models exhibit substantial pose errors, and fine-tuning under the shared wrist-camera formulation (Sec.~\ref{sec:expert_training}) reduces both translation and rotation errors. These results highlight the importance of fine-tuning pretrained 3D models on robot data for wrist-camera motion estimation. Fine-tuned VGGT achieves the lowest translation RMSE, rotation RMSE, and gripper MAE, making it our default backbone.

\paragraph{Pretraining and pose supervision.}

\begin{figure}[t]
    \centering
    \includegraphics[width=\linewidth]{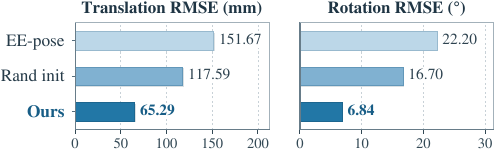}
    \caption{\textbf{Effect of pretraining and pose supervision on Galaxea pose estimation.} Rand init replaces pretrained initialization with random weights; EE-pose replaces wrist-camera supervision with end-effector poses.}
    \label{fig:pose-ablation}
\end{figure}

We investigate two design choices for training the VGGT expert: pretrained versus random initialization, and wrist-camera versus end-effector poses as supervision targets. As shown in Fig.~\ref{fig:pose-ablation}, random initialization leads to substantially higher translation and rotation errors than pretrained initialization, supporting the use of pretrained 3D priors for wrist-camera motion estimation. Supervising the model with wrist-camera poses also yields more accurate trajectories than using end-effector poses. Wrist-camera supervision retains the camera-pose estimation task used in VGGT pretraining and provides a common geometric target across all five datasets. This formulation avoids learning embodiment-specific calibration. Together, these results support combining pretrained 3D priors with wrist-camera supervision to learn accurate motion estimation from heterogeneous robot data.

\paragraph{Effect of Model Stitching.}

To assess how model stitching affects action recovery, we compare trajectories from the same video prediction using the trained RGB expert on decoded frames and the stitched estimator on predicted latents. The predicted wrist views show an approach toward the lettuce~(Fig.~\ref{fig:stitch-ablation}a). The RGB expert recovers a mostly forward trajectory with insufficient descent, whereas the stitched estimator captures the downward motion needed to reach the lettuce~(Fig.~\ref{fig:stitch-ablation}b). Executing the stitched trajectory brings the gripper to the lettuce, with wrist views closely matching the predicted approach~(Fig.~\ref{fig:stitch-ablation}a, bottom). On PiPER-X, the stitched estimator achieves 40\% success on \textit{put the vegetable on the bread}, compared with 0\% for the RGB expert~(Fig.~\ref{fig:stitch-ablation}b). Together, these results show that a plausible visual plan can be poorly translated by RGB-based pose estimation. By recovering motion more consistent with the visual plan and improving task execution, model stitching provides an effective interface between video prediction and robot control without intermediate RGB decoding.

%% file: sec/6_conclusion.tex
\section{Conclusion}

We presented \methodname{}~(\methodnameshort{}), a policy that decouples task prediction from motion execution through wrist-camera motion. A video model predicts how a task unfolds, while a 3D pose estimator translates those predictions into actions through a shared geometric formulation. Trained on public robot data, the estimator transfers across three platforms without further fine-tuning. Comparisons with VERA and Cosmos3 show higher task success with our pose estimator and gains from switching video generators. This separation offers a scalable path to generalist manipulation: task knowledge can expand through large-scale video learning while geometric motion estimation remains reusable across tasks and embodiments, allowing advances in video prediction to improve robot control without new action-labeled demonstrations for real-world deployment.

% We showed that estimating wrist-camera motion provides a direct way to turn video predictions into robot actions. This formulation lets a video model predict how a task unfolds while a 3D pose estimator solves the same geometric problem across tasks and embodiments. In real-world experiments, the pose estimator trained on public robot data transferred across three platforms without further fine-tuning. These results show how wrist-camera motion can extend the benefits of video pretraining to robot control while reducing the need to collect new action labels for each deployment.

%% file: sec/X_suppl.tex
\clearpage
\maketitlesupplementary

% \section{Rationale}
% \label{sec:rationale}
% % 
% Having the supplementary compiled together with the main paper means that:
% % 
% \begin{itemize}
% \item The supplementary can back-reference sections of the main paper, for example, we can refer to \cref{sec:intro};
% \item The main paper can forward reference sub-sections within the supplementary explicitly (e.g. referring to a particular experiment); 
% \item When submitted to arXiv, the supplementary will already included at the end of the paper.
% \end{itemize}
% % 
% To split the supplementary pages from the main paper, you can use \href{https://support.apple.com/en-ca/guide/preview/prvw11793/mac#:~:text=Delete%20a%20page%20from%20a,or%20choose%20Edit%20%3E%20Delete).}{Preview (on macOS)}, \href{https://www.adobe.com/acrobat/how-to/delete-pages-from-pdf.html#:~:text=Choose%20%E2%80%9CTools%E2%80%9D%20%3E%20%E2%80%9COrganize,or%20pages%20from%20the%20file.}{Adobe Acrobat} (on all OSs), as well as \href{https://superuser.com/questions/517986/is-it-possible-to-delete-some-pages-of-a-pdf-document}{command line tools}.

\section{Dataset Details}

To adapt the pose estimator to robotic manipulation, we construct training pairs of robot videos and wrist-camera trajectories from AgiBot~\cite{contributors2025agibot}, DROID~\cite{khazatsky2024droid}, BEHAVIOR-1K~\cite{li2024behaviork}, InternData-A1~\cite{tian2025interndataa}, and RH20T~\cite{fang2023rht}, using the subsets summarized in Tab.~\ref{tab:data_sources}.
We select these datasets because they provide wrist-camera observations whose poses are either recorded directly or can be recovered from robot states and camera calibrations. We exclude episodes whose wrist-camera poses cannot be obtained. DROID provides wrist-camera poses in the robot-base frame, which we use directly. For AgiBot and RH20T, we convert the recorded end-effector or TCP poses into camera poses using the corresponding calibration transforms. InternData-A1 and BEHAVIOR-1K include recordings with moving bases, where wrist-camera poses are expressed in the moving robot-base frame. For InternData-A1, we use the recorded base poses to align the camera trajectory to a fixed reference frame.
BEHAVIOR-1K records timestamped planar base velocities, including local translational velocities and yaw rates. We integrate these measurements to recover the base poses needed for the alignment, using the matrix-exponential update
\begin{equation}
U_{j+1}
=
U_j\exp\!\left(\Delta t_j\,\widehat{\bar{\xi}}_j\right),
\label{eq:base_integration}
\end{equation}
where $U_j$ denotes the base pose at original frame $j$ relative to the window's initial base frame, with $U_0=I$. Here, $\Delta t_j$ is the interval between consecutive timestamps, $\bar{\xi}_j$ is the mean of the body-frame twists at the two interval endpoints, and the hat operator maps the twist vector to its matrix representation.

\begin{table*}[t]
    \centering
    \caption{\textbf{Data sources for multi-source trajectory supervision.}
    We combine real and simulated robot datasets spanning diverse embodiments and manipulation environments. Statistics correspond to the selected subsets with wrist-camera observations and camera poses available directly or recoverable from robot states and calibrations.}
    \label{tab:data_sources}
    \small
    \setlength{\tabcolsep}{5pt}
    \renewcommand{\arraystretch}{1.2}
    \begin{tabularx}{\textwidth}{
        @{}p{0.16\textwidth} X c X r r r@{}
    }
        \toprule
        \textbf{Dataset} &
        \textbf{Embodiment} &
        \textbf{Domain} &
        \textbf{Scene} &
        \textbf{\# Episodes} &
        \textbf{\# Frames} &
        \textbf{\# Hours} \\
        \midrule

        AgiBot World\newline
        (calibrated subset) &
        AgiBot G1\newline
        (dual-arm) &
        Real &
        Tabletop, household, and retail environments &
        17,092 &
        28,893,564 &
        267.53 \\

        \addlinespace[3pt]
        DROID\newline
        (success + failure) &
        Franka Panda\newline
        (single-arm) &
        Real &
        Tabletop, drawers, cabinets, and kitchens &
        66,411 &
        17,619,832 &
        326.29 \\

        \addlinespace[3pt]
        BEHAVIOR-1K &
        R1Pro\newline
        (mobile dual-arm) &
        Sim. &
        Indoor environments with mobile manipulation &
        20,000 &
        210,916,774 &
        1952.93 \\

        \addlinespace[3pt]
        InternData-A1\newline
        (\texttt{sim\_updated}) &
        Franka, ARX Lift2,\newline
        Split Aloha, and Genie1\newline
        (single-/dual-arm) &
        Sim. &
        Tabletop, shelves, and household appliances &
        586,631 &
        394,138,291 &
        3649.43 \\

        \addlinespace[3pt]
        RH20T\newline
        (cfg1--7 subset) &
        Flexiv, UR5, Franka Panda,\newline
        and KUKA (single-arm) &
        Real &
        Workbenches with tools and fixtures &
        2,875$^{\dagger}$ &
        1,010,021 &
        31.12 \\

        \addlinespace[3pt]
        \textbf{Total} &
        - &
        - &
        - &
        693,009 &
        652,578,482 &
        6227.31 \\

        \bottomrule
    \end{tabularx}

    \smallskip
    \parbox{\textwidth}{\footnotesize
        $^{\dagger}$RH20T contains 3,069 candidate episodes in the
        selected configurations. The reported episode and frame counts
        cover the 2,875 episodes with readable wrist-camera timestamps.
    }
\end{table*}

\section{Training Details}

\label{app:training_details}

We train the RGB pose estimator and its stitched variants using the five public datasets described in Tab.~\ref{tab:data_sources}. The training configuration is summarized in Tab.~\ref{tab:training_config}. Both stages use the same robot videos, with dataset annotations supervising the RGB expert and the frozen expert providing targets for distillation.

\paragraph{Training samples.}
We sample 17-frame clips at a randomly selected observation frequency between 2 and 10\,Hz. The RGB expert takes five anchor frames at indices $\{0,4,8,12,16\}$, each resized to $224\times280$ pixels. For stitching, the frozen Wan VAE~\cite{wan2025wan} encodes the full clip with fourfold temporal compression. The first frame produces the initial latent, and every four subsequent frames produce one additional latent time step, yielding five latent time steps from 17 RGB frames. We align the pose and gripper targets with the five anchor frames. VAE preprocessing follows the corresponding video generator. For VERA~\cite{lester2026turning}, each view is resized to $128\times192$ and encoded separately. For VPP2~\cite{yanjiang2026video}, the views are tiled into a single $240\times416$ image before VAE encoding. All image dimensions are reported as height $\times$ width.

\paragraph{RGB expert training.}
We initialize the backbone and camera head from the official pretrained VGGT~\cite{jianyuanvggt} weights and jointly optimize them with the gripper head. Camera supervision uses VGGT's multi-stage $L_1$ loss on translation and rotation, with equal weights for the two terms and a stage decay factor of $0.6$. Focal-length supervision is disabled. The gripper head predicts continuous gripper opening normalized to $[0,1]$ and is supervised with a smooth $L_1$ loss. The camera and gripper losses receive weights of $5.0$ and $0.1$, respectively. We use the checkpoint obtained after 50,000 minibatch iterations.

\paragraph{Model stitching and distillation.}
After probing, the fifth block of the VGGT backbone yields the least error, because of which we select it as our stitch point. 
The video latents are connected to the expert's image-encoder features using a $1\times1$ convolutional adapter with 1,024 output channels. The retained backbone and prediction heads are initialized from the RGB expert. During distillation, the expert processes the original wrist-camera frames and the student processes their corresponding VAE latents. The expert and VAE remain frozen, while the adapter, retained backbone, and prediction heads are jointly optimized for 10,000 minibatch iterations. Camera predictions are distilled using the same multi-stage pose loss, and the student's sigmoid gripper predictions are matched to the expert's predicted gripper openings using a smooth $L_1$ loss. We also apply latent noise augmentation with a maximum noise standard deviation of $0.2$. The student uses BF16 mixed precision, while the frozen expert runs in FP32.

\begin{table*}[t]
    \centering
    \small
    \caption{\textbf{Training configurations of the RGB expert and stitched pose estimator.} 
    The RGB expert learns wrist-camera motion and gripper opening from dataset supervision. The stitched estimator learns to reproduce the frozen expert's predictions from VAE-encoded video latents. The stitching configuration applies to the expert-distilled VERA and VPP2 variants.}
    \label{tab:training_config}
    \setlength{\tabcolsep}{12pt}
    \renewcommand{\arraystretch}{1.12}
    \begin{tabular}{lcc}
        \toprule
        \textbf{Setting} &
        \textbf{RGB expert} &
        \textbf{Stitched estimator} \\
        \midrule
        Initialization & Pretrained VGGT & Trained RGB expert \\
        Supervision & Dataset annotations & Frozen expert predictions \\
        Optimizer & AdamW & AdamW \\
        Backbone learning rate & $10^{-6}$ & $10^{-5}$ \\
        Head learning rate & $10^{-5}$ & $10^{-5}$ \\
        Adapter learning rate & -- & $10^{-5}$ \\
        Weight decay & $10^{-4}$ & $10^{-4}$ \\
        Learning-rate schedule & Cosine & Cosine \\
        Minimum learning-rate ratio & $0.1$ & $0.1$ \\
        Warmup updates & 1,000 & 100 \\
        Gradient clipping norm & $1.0$ & $1.0$ \\
        Global batch size & 128 & 128 \\
        Training updates & 25,000 & 5,000 \\
        Gradient accumulation steps & 2 & 2 \\
        Training precision & BF16 & BF16 \\
        Reference RGB resolution & $224\times280$ & $224\times280$ \\
        Camera loss & Multi-stage $L_1$ & Multi-stage $L_1$ \\
        Camera loss weight & $5.0$ & $5.0$ \\
        Stage decay factor & $0.6$ & $0.6$ \\
        Gripper loss & Smooth $L_1$ & Smooth $L_1$ \\
        Gripper loss weight & $0.1$ & $0.1$ \\
        Random seed & 42 & 42 \\
        \bottomrule
    \end{tabular}
\end{table*}

\section{Inference Details}

\begin{table*}[t]
    \centering
    \small
    \caption{\textbf{Task instructions and execution horizons.}
    For each task, we list the exact instruction provided to the policy and the number of actions executed before replanning.}
    \label{tab:inference_configs}
    \renewcommand{\arraystretch}{1.15}
    \begin{tabularx}{\textwidth}{@{}lXc@{}}
        \toprule
        Embodiment & Prompt & Execution steps \\
        \midrule
        Galaxea R1 Pro
        & open the washing machine door & 8 \\
        Galaxea R1 Pro
        & grip the knob with the gripper and maintain the grasping posture & 8 \\
        Galaxea R1 Pro
        & pick up the fallen clothes from the floor & 8 \\
        Galaxea R1 Pro
        & take the clothes out of the washing machine & 8 \\
        \midrule
        Franka Panda
        & pick the pen into the box & 4 \\
        Franka Panda
        & stack the bowls & 4 \\
        Franka Panda
        & put the basketball in the basket & 6 \\
        Franka Panda
        & open the drawer & 8 \\
        \midrule
        PiPER-X
        & insert the bread into the toaster, then release the bread & 8 \\
        PiPER-X
        & use the towel to wipe the plate & 8 \\
        PiPER-X
        & put the vegetable on the bread & 6 \\
        PiPER-X
        & put the rubik's cube on the plate & 4 \\
        \bottomrule
    \end{tabularx}
\end{table*}

We list our specific prompts and action execution steps in Tab.~\ref{tab:inference_configs}. At each planning step, the policy predicts wrist-camera pose increments and gripper states. We convert these increments into end-effector targets using hand-eye calibration and execute the specified number of actions before replanning. 

\paragraph{Galaxea R1 Pro.}
We accumulate the predicted motion increments from the measured end-effector poses to obtain target trajectories for both arms. Positions are linearly interpolated, while orientations are interpolated using spherical linear interpolation. RelaxedIK converts the interpolated targets into joint configurations for each arm. The mobile base remains stationary during execution.

\paragraph{Franka and PiPER-X.}
We convert the predicted wrist-camera motion into arm targets and compute joint configurations using numerical inverse kinematics. We use Pinocchio for kinematic computations, with bounded nonlinear least-squares optimization for Franka and damped least-squares iterations for PiPER-X. The resulting joint targets are interpolated with quintic polynomials and executed together with the predicted gripper commands.

\section{Additional Analyses}

\paragraph{Model Stitching.}

\begin{figure}[t]
    \centering
    \includegraphics[width=\linewidth]{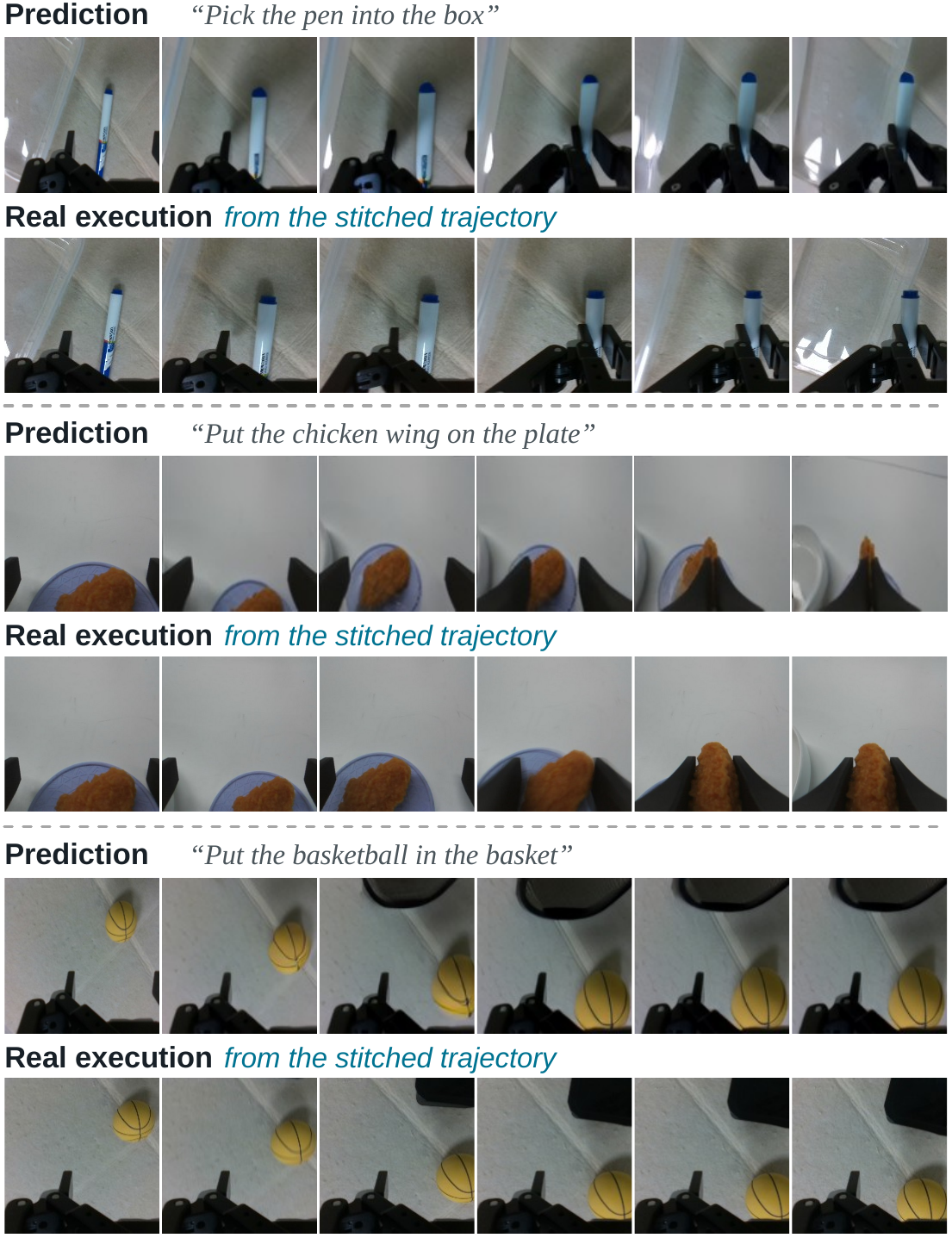}
    \caption{\textbf{Model stitching visualization.}
    For each task, the upper row shows decoded video predictions, and the lower row shows wrist-camera observations recorded during execution.}
    \label{fig:stitch-visualization}
\end{figure}

\begin{table}[t]
    \centering
    \small
    \caption{\textbf{Estimation accuracy before and after model stitching.}
    Both models are evaluated on the Galaxea benchmark. Translation and rotation are measured by RMSE, and normalized gripper opening by MAE.}
    \label{tab:stitch_accuracy}
    \setlength{\tabcolsep}{4pt}
    \begin{tabular}{@{}llccc@{}}
        \toprule
        Model & Input
        & \shortstack{Trans.\\(mm) $\downarrow$}
        & \shortstack{Rot.\\($^\circ$) $\downarrow$}
        & \shortstack{Gripper\\MAE $\downarrow$} \\
        \midrule
        RGB expert & RGB
        & \textbf{65.29} & \textbf{6.84} & \textbf{0.203} \\
        Stitched VGGT & Latents
        & 113.31 & 15.94 & 0.271 \\
        \bottomrule
    \end{tabular}
\end{table}

We compare the RGB expert and stitched VGGT on recorded videos from the same Galaxea benchmark. As shown in Tab.~\ref{tab:stitch_accuracy}, the RGB expert achieves lower errors across all three metrics. The execution gains from stitching therefore cannot be attributed to better estimation accuracy on recorded videos. Fig.~\ref{fig:stitch-visualization} further compares the generated wrist views with observations recorded during execution of the stitched trajectories. Across the three examples, the robot closely follows the predicted approach and grasping motions, even when the generated images contain blur or distorted object shapes. For example, in \textit{put the chicken wing on the plate}, the generated chicken wing becomes distorted as the gripper closes, while the robot executes the corresponding grasp on the real object. In \textit{put the basketball in the basket}, the predicted basket has a visibly different contour from the real one, yet the executed motion follows the approach shown in the prediction. Together with the offline comparison, these examples support the benefit of model stitching, which allows the pose estimator to recover executable motion directly from predicted latents despite visible artifacts in the decoded video.

\paragraph{Action execution steps.}

\begin{figure}[t]
    \centering
    \includegraphics[width=\linewidth]{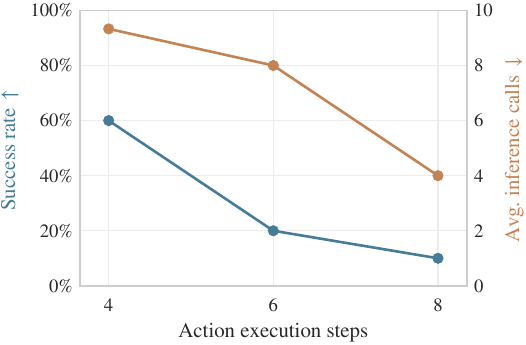}
    \caption{\textbf{Effect of the action execution horizon.} Success rate is measured over 10 trials. We also report the average number of inference calls for each execution horizon.}
    \label{fig:action-execution}
\end{figure}

We vary the execution horizon on \textit{pick the pen into the box} using \methodnameshort{} with VERA-DROID on Franka~(Fig.~\ref{fig:action-execution}). Increasing the horizon from 4 to 8 reduces the average inference calls among successful trials from 9.33 to 4, but lowers success from 60\% to 10\%. More frequent replanning allows the policy to update its predictions and correct execution deviations sooner. These results illustrate the trade-off between execution success and inference cost. Execution horizons for all 12 tasks are listed in Tab.~\ref{tab:inference_configs}.

\paragraph{Inference time.}

\begin{table}[t]
    \centering
    \small
    \caption{\textbf{Inference latency under the evaluated configurations.}
    Values are means over three runs. VERA uses two views with our stitched estimator and three views with its Jacobian IDM, matching their respective real-world inference configurations. VERA and VPP2 use 40 and 30 denoising steps, respectively. All timings are measured on a single NVIDIA A800 80GB GPU.}
    \label{tab:inference_time}
    \begin{tabular}{@{}lr@{}}
        \toprule
        Model / stage & Time (s) \\
        \midrule
        $\pi_{0.5}$-DROID & 0.092 \\
        Cosmos3-Nano-DROID (with video decoding) & 3.420 \\
        VERA + Jacobian IDM (full call) & 56.825 \\
        \midrule
        \textbf{\methodnameshort{} with VERA (full call)} & 38.947 \\
        \quad Video generation & 38.740 \\
        \quad Stitched pose estimation & 0.193 \\
        \addlinespace
        \textbf{\methodnameshort{} with VPP2 (full call)} & 53.527 \\
        \quad Video generation & 53.351 \\
        \quad Stitched pose estimation & 0.174 \\
        \bottomrule
    \end{tabular}
\end{table}

Tab.~\ref{tab:inference_time} reports inference latency measured using observations collected during real-world execution on Franka, averaged over three runs. Video generation dominates the latency, while our stitched pose estimator takes less than 0.2\,s on average. These measurements use multi-step video generation. One-step distillation, as explored in Dyna-2 and VPP2~\cite{robotics2026dyna,yanjiang2026video}, offers a promising direction for acceleration, which we leave to future work.

\section{Details on Video Generator Evaluation}

\label{app:video_generators}

\paragraph{Benchmark and generator inputs.}

\begin{table}[t]
    \centering
    \small
    \caption{Composition of the PiPER wrist-camera benchmark.}
    \label{tab:video_benchmark_tasks}
    \begin{tabular}{@{}lr@{\qquad}lr@{}}
        \toprule
        Task & Clips & Task & Clips \\
        \midrule
        Folding clothes   & 23 & Inserting a hanger & 22 \\
        Handling a towel  & 22 & Placing a pen      & 10 \\
        Wiping a table    & 22 & Pushing an object  & 22 \\
        Tidying a table   & 22 & Grasping a block   & 23 \\
        Pouring water     & 23 & Bagging items      & 22 \\
        Arranging flowers & 22 & Stacking dishes    & 23 \\
        \midrule
        \multicolumn{3}{l}{Total} & 256 \\
        \bottomrule
    \end{tabular}
\end{table}

We compare six generators on 256 clips spanning 12 manipulation tasks collected with PiPER, with the task distribution summarized in Tab.~\ref{tab:video_benchmark_tasks}. 

All six models receive the same instruction and initial global and wrist views. They generate the three views jointly, using the input format appropriate to each generator. VERA-OMNI~\cite{lester2026turning} encodes each view separately and arranges the latents in its native multi-view format. VPP2~\cite{yanjiang2026video} arranges the input views into a single image. For Wan2.1~\cite{wan2025wan} and LVP~\cite{chen2025large}, we encode the three views separately and concatenate the latents along their width. For Cosmos Predict2.5~\cite{arslan2025world}, we likewise encode each view with its Wan2.1 VAE and concatenate the latents along their width. For Cosmos3~\cite{agarwal2026cosmos}, we first arrange the RGB images on a canvas with the global view at the upper left, the two wrists along the bottom, and the remaining tile blank. We then encode the full canvas and separate the views after generation.

\paragraph{Qualitative comparison.}

\begin{figure*}[htbp]
    \centering
    \includegraphics[width=\textwidth]{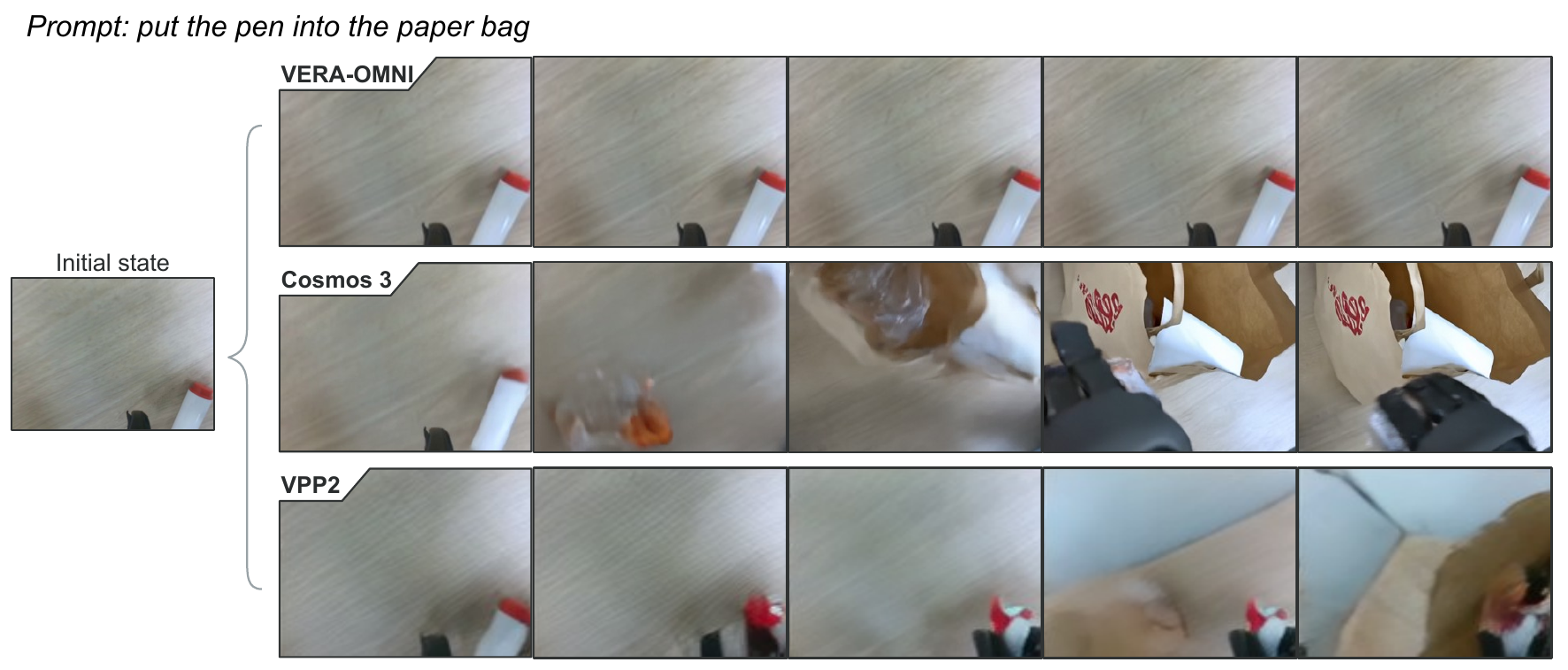}
    \caption{\textbf{Qualitative comparison of wrist-view predictions.} Given the same initial observations and instruction, the generators produce different levels of visual consistency and task progression. Each row shows selected frames from one wrist view in temporal order.}
    \label{fig:app_generator_comparison}
\end{figure*}

Fig.~\ref{fig:app_generator_comparison} illustrates the distinction between physical adherence and task progression for the instruction \textit{put the pen into the paper bag}. VERA-OMNI preserves the appearance of the pen and gripper but predicts little movement toward the goal. Cosmos3 produces larger scene changes, but the pen becomes difficult to track through the intermediate frames, where an orange shape appears on the tabletop. VPP2 shows the pen remaining near the gripper as the view moves toward the bag opening, combining greater visual consistency with clearer task progression.

\paragraph{Evaluation protocol.}

We use RoboJudge-9B~\cite{siyuan2026robojudge} to evaluate Physical Adherence (PA) and Instruction Alignment (IA). PA measures whether the robot and objects remain consistent over time and whether motion and contact are physically plausible. IA measures whether the intended manipulator acts on the correct object and how much of the requested task is visibly completed.

\begin{figure*}[p]
    \centering
    \includegraphics[width=\textwidth]{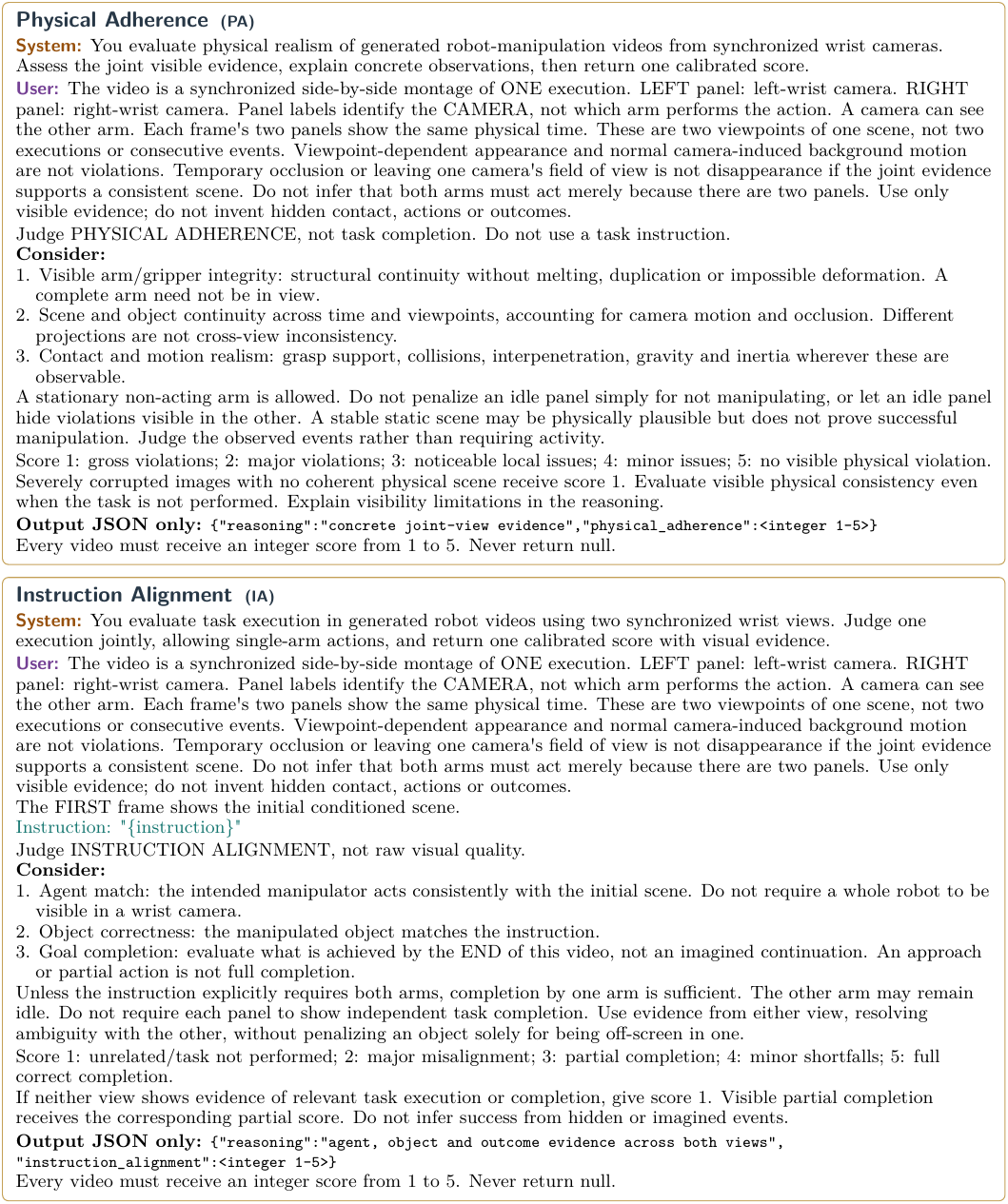}
    \caption{\textbf{RoboJudge prompts for evaluating dual-wrist videos.} The top box shows the prompt for Physical Adherence and the bottom box shows the prompt for Instruction Alignment. The placeholder \texttt{\{instruction\}} is replaced with the task instruction for the clip.}
    \label{fig:robojudge}
\end{figure*}

We concatenate the synchronized wrist views side by side and label each camera, prompting the judge to assess them as complementary views of a single execution. This allows an object to leave one view while remaining visible in the other, and one arm to complete the task while the other remains still. We make two calls per clip: PA is evaluated without the instruction, while IA additionally receives the instruction and uses the first frame as the initial scene. Each call returns an integer score from 1 to 5 with supporting visual evidence. We report mean PA and IA separately over all 256 clips. The evaluation prompts are reproduced in Fig.~\ref{fig:robojudge}.

\paragraph{Pose estimator pairing.}

RoboJudge scores only the generated video. To evaluate execution, we train a corresponding pose estimator for each of the six generators. We pair each generator with a pose estimator trained in its corresponding VAE latent space. VERA-OMNI, VPP2, Wan2.1, LVP, and Cosmos Predict2.5 use the Wan2.1 VAE, while Cosmos3 uses the Wan2.2 VAE. For each pairing, training inputs follow the generator's view arrangement and VAE preprocessing described above, so the estimator receives the same latent format during training and inference.

\section{Details on Video Model Fine-tuning}

\label{app:video_finetuning}

We collect synchronized multi-view RGB videos with Galaxea R1 Pro in a laundry-room setting, covering washing-machine door operation, loading and retrieving clothes, and handling laundry baskets and bags.

We fine-tune the video generator following VERA's training procedure~\cite{lester2026turning}, initializing from its released VERA-OMNI checkpoint based on Wan2.1-I2V-14B. Each training sample contains 37 frames sampled with a stride of two, corresponding to 15\,Hz for recordings at 30\,fps. The inputs comprise two head-camera views and two wrist-camera views, each resized to $128\times192$. We encode each view separately with the frozen VAE and concatenate the resulting latents. Following Diffusion Forcing, we randomly vary the history length from one to six latent time steps and predict four future latent steps, conditioned on the task instruction, CLIP image features, and historical latents. The DiT is trained with an MSE flow-matching loss whose target is the difference between the sampled noise and the clean video latents.

We fine-tune all DiT parameters while freezing the VAE, text encoder, and CLIP image encoder. Training runs for 6,000 optimizer steps on eight GPUs using FSDP and BF16 precision, with a batch size of 20 per GPU and a global batch size of 160. We use AdamW with a learning rate of $10^{-5}$, betas of $(0.9,0.95)$, and weight decay of $0.05$. The learning rate warms up over the first 1,000 steps and remains constant thereafter.

\section{Details on Pose Estimator Evaluation}

\label{app:pose_estimators}

\paragraph{Benchmark construction.}
We construct an offline pose estimation benchmark from 1,000 Galaxea R1 Pro episodes sampled from the dataset discussed in Sec.~\ref{app:video_finetuning}. The laundry-room environment is absent from the five public datasets used to train the pose estimators. Each episode provides synchronized wrist-camera images, recorded end-effector poses, gripper states, and hand-eye calibrations, allowing us to evaluate motion estimation in an unseen environment.
For each episode, we fix a randomly selected starting frame and sample five synchronized pairs of wrist-camera images at offsets of $\{0,24,48,72,96\}$ frames. At 30\,fps, this corresponds to sampling at 5\,Hz with an anchor stride of four, consistent with our training setup. All models receive the same images at the same time points, resized to their respective input resolutions.

\paragraph{Evaluation protocol.}
We evaluate all models in end-effector pose space. Camera-pose predictions are converted into end-effector trajectories using the corresponding hand-eye calibration. The VGGT variant trained with end-effector pose supervision is evaluated directly from its predicted trajectories. Both predicted and reference trajectories are expressed relative to their respective initial end-effector poses, and we score the four subsequent poses.

Each wrist is evaluated separately, yielding 4,000 evaluation points per wrist when all trajectories are valid. We report end-effector translation RMSE in millimeters, rotation RMSE in degrees, and gripper MAE for models equipped with a gripper head. We pool the evaluation points from both wrists to compute the reported metrics.

\section{Limitations and Future Work}

\paragraph{Inference efficiency.}

The inference latency of video models such as VERA and VPP2 limits the efficiency of real-world deployment, as the robot must wait for new predictions between action segments~(Tab.~\ref{tab:inference_time}). Executing longer action segments reduces the number of inference calls but also limits opportunities for feedback correction, as shown in Fig.~\ref{fig:action-execution}. Accelerating the video generator through one-step distillation~\cite{robotics2026dyna,yanjiang2026video} could enable more frequent replanning while reducing waiting time.

\paragraph{Wrist-camera occlusion.}

Motion estimation becomes challenging when a manipulated object occupies most of the wrist-camera view. For example, when handling a large bowl, the camera may observe mainly the bowl rather than the surrounding scene. As the bowl moves together with the gripper, its appearance provides limited evidence of camera motion relative to the environment. Incorporating auxiliary views into the pose estimator could provide complementary geometric cues and improve robustness to such occlusions.

\paragraph{Video prediction accuracy.}

Inaccuracies in video prediction are a major factor limiting execution success. Despite their broad generalization capabilities, video generators can still predict slightly incorrect viewpoints or spatial relationships, such as the gripper's height relative to an object. These deviations can lead to unsuccessful execution even when the pose estimator faithfully recovers the predicted motion. Improving the geometric accuracy of video predictions is therefore an important direction for increasing manipulation reliability.

\paragraph{More demanding manipulation.}

Our evaluation does not yet cover highly precise operations such as screw driving, more complex bimanual coordination, or tasks involving large rotations. Pose estimation errors may become more consequential in these settings, where successful execution requires tighter positional and rotational accuracy. Future work could explore these tasks and further investigate few-shot adaptation of the video generator and pose estimator to improve precision and coordination with additional fine-tuning data.

\section{Additional Visualization}

\begin{figure*}[p]
    \centering
    \includegraphics[width=\linewidth]{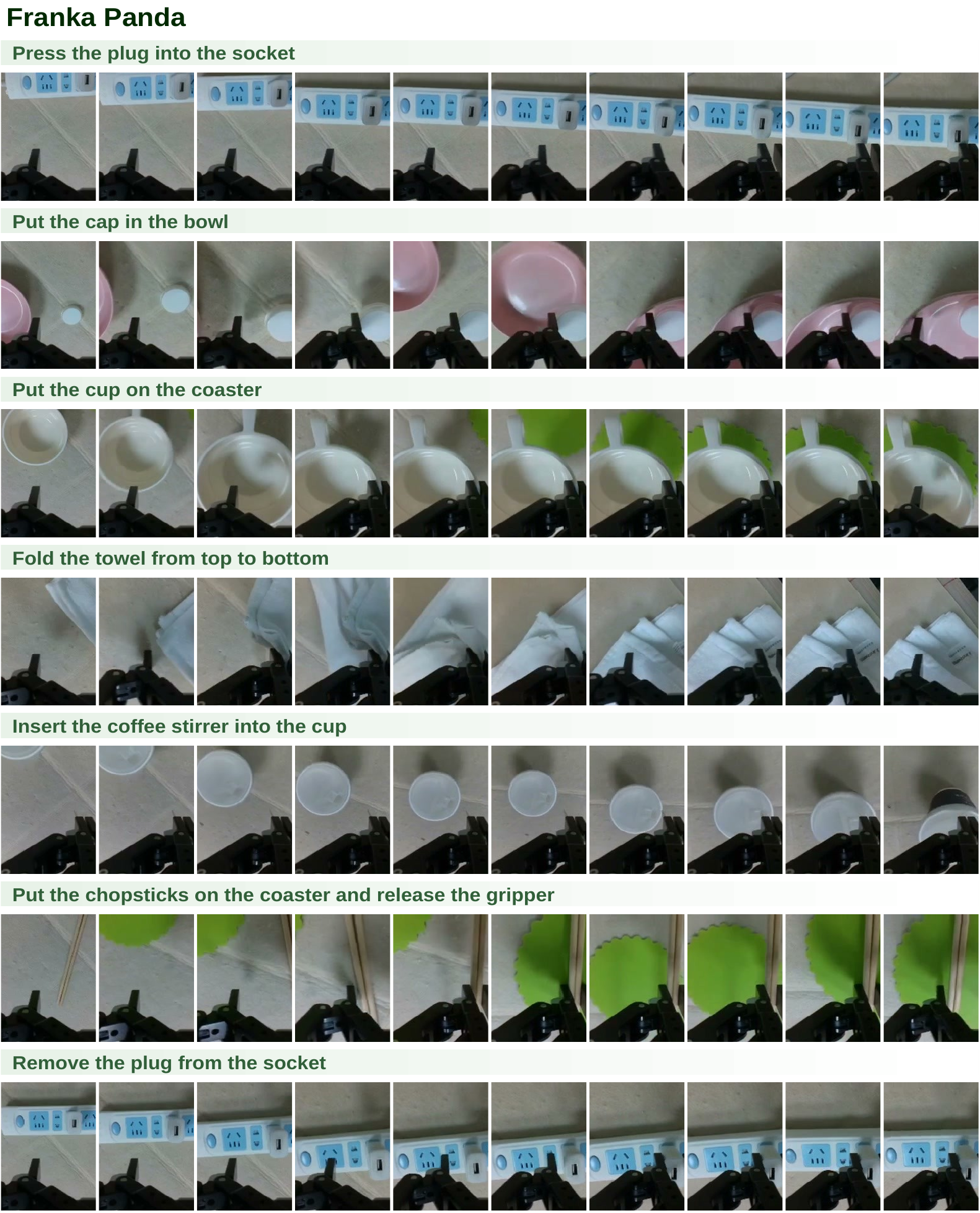}
    \caption{\textbf{Additional real-world executions on Franka.}}
    \label{fig:franka}
\end{figure*}

\begin{figure*}[p]
    \centering
    \includegraphics[width=\linewidth]{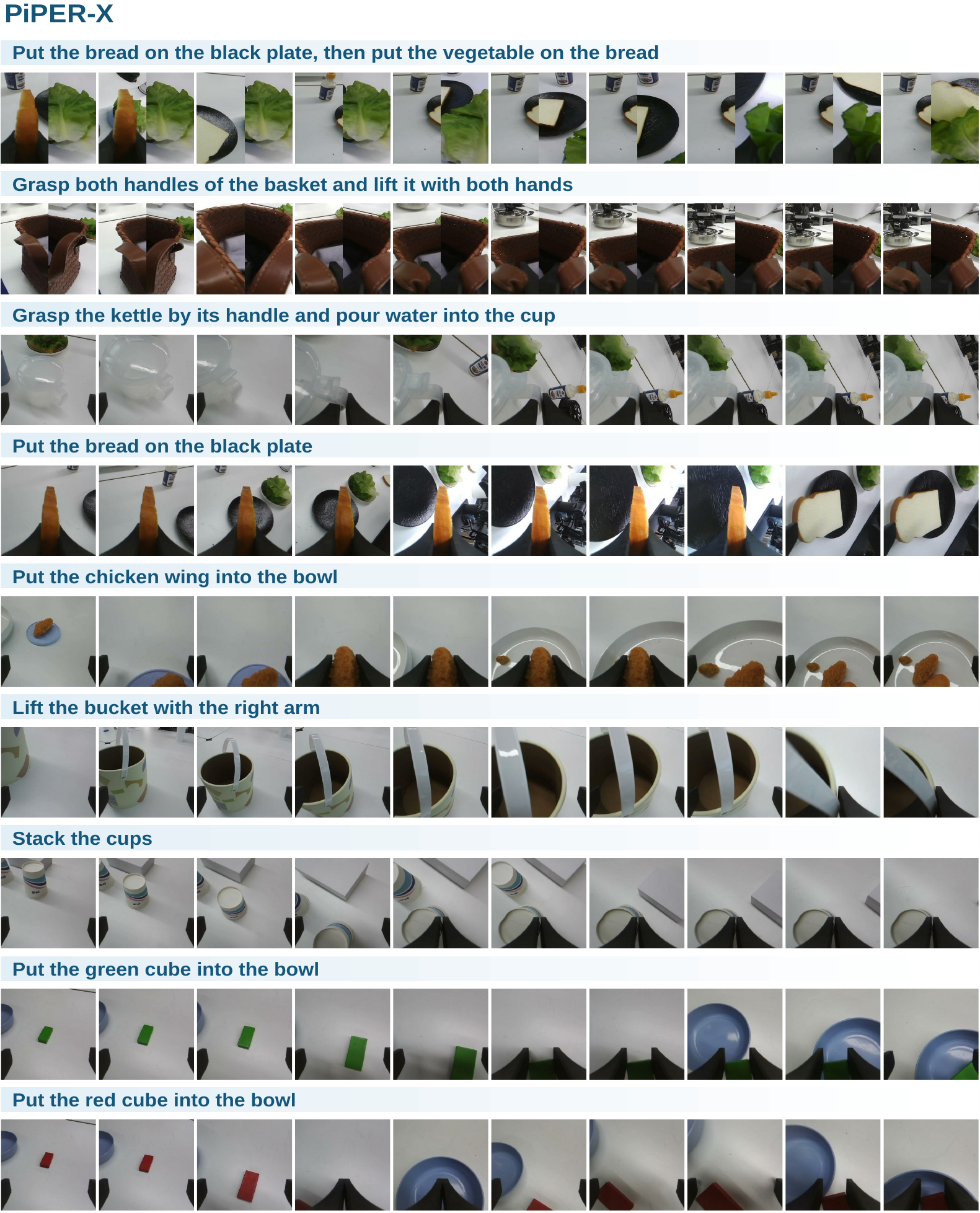}
    \caption{\textbf{Additional real-world executions on PiPER-X.} For the first two tasks, the left and right wrist views are concatenated.}
    \label{fig:piper}
\end{figure*}

Fig.~\ref{fig:franka} and Fig.~\ref{fig:piper} present additional real-world executions on Franka and PiPER-X, extending the task and object coverage shown in the main paper. These examples further illustrate the following capabilities:
\begin{itemize}
    \item \textbf{Diverse manipulation skills.}
    The tasks extend beyond pick-and-place to towel folding, plug insertion and removal, and tool use such as pouring water from a kettle. 
    
    \item \textbf{Fine-grained manipulation of varied objects.}
    The examples cover objects with different shapes, sizes, and materials, including cups, food items, and towels. Small objects, such as bottle caps, coffee stirrers, and chopsticks, further require precise manipulation.
    
    \item \textbf{Bimanual coordination.}
    The two arms perform complementary actions when placing bread on a plate and subsequently placing a vegetable on the bread. Lifting a basket by both handles additionally requires the arms to coordinate their grasps and lifting motions.
\end{itemize}

